\documentclass[sigconf]{acmart}

\usepackage{algorithm}
\usepackage{algorithmic}
\usepackage{diagbox}
\usepackage{makecell}
\usepackage{subfigure}
\usepackage{enumitem}
\usepackage[frozencache,cachedir=.]{minted}
\usepackage{newtxmath}
\usepackage{multirow}
\usepackage{svg}
\usepackage{caption}
\AtBeginDocument{%
  \providecommand\BibTeX{{%
    \normalfont B\kern-0.5em{\scshape i\kern-0.25em b}\kern-0.8em\TeX}}}

\usepackage{balance}

\newcommand{\ours}{SageCopilot}
\newcommand{\DuSQL}{DuSQL}
\newcommand{\industrial}{Real Traffic Dataset}

\begin{document}

\title[\ours{}]{Towards Automated Data Sciences with Natural Language and \ours{}: Practices and Lessons Learned}
\author{Yuan Liao$^1$, Jiang Bian$^1$, Yuhui Yun$^1$, Shuo Wang$^1$, Yubo Zhang$^{1,2}$, Jiaming Chu$^{1,3}$, \\Tao Wang$^{1,3}$, Kewei Li$^1$, Yuchen Li$^1$, Xuhong Li$^1$, Shilei Ji$^1$, Haoyi Xiong$^1$}
\affiliation{
    \institution{$^1$Baidu Inc., $^2$Beihang University, $^3$Beijing University of Posts and Telecommunications}
    \city{Haidian District, Beijing}
    \country{China}
}







\renewcommand{\shortauthors}{Yuan Liao et al.}

\begin{abstract}
While the field of NL2SQL has made significant advancements in translating natural language instructions into executable SQL scripts for data querying and processing, achieving full automation within the broader data science pipeline--encompassing data querying, analysis, visualization, and reporting--remains a complex challenge. This study introduces \ours{}, an advanced, industry-grade system system that automates the data science pipeline by integrating Large Language Models (LLMs), Autonomous Agents (AutoAgents), and Language User Interfaces (LUIs). Specifically, \ours{} incorporates a two-phase design: an online component refining users' inputs into executable scripts through In-Context Learning~\cite{dong2022survey} (ICL) and running the scripts for results reporting \& visualization, and an offline preparing demonstrations requested by ICL in the online phase. A list of trending strategies such as Chain-of-Thought and prompt-tuning have been used to augment \ours{} for enhanced performance. Through rigorous testing and comparative analysis against prompt-based solutions, \ours{} has been empirically validated to achieve superior end-to-end performance in generating/executing scripts and offering results with visualization, backed by real-world datasets. Our in-depth ablation studies highlight the individual contributions of various components and strategies used by \ours{} to the end-to-end correctness for data sciences.



\end{abstract}

\ccsdesc[500]{Information systems~Computing platforms}

\keywords{Large Language Models, AutoAgents, Data Analysis, In-Context Learning, Human-Computer Interaction.}

\maketitle

\section{Introduction}\label{introduction}
The advancements in Text-to-SQL generation propose promising approaches to convert textual instructions into structured query language (SQL) for database operations \cite{zhongSeq2SQL2017,gualtieri2017interfaces,xuSQLNet2017,wang2019rat,wang2020dusql}. However, achieving full automation in the data science pipeline is challenging \cite{gualtieri2017interfaces} due to the complexity of understanding user intents \cite{islam2012modeling,gan2021natural}, integrating access control mechanisms \cite{hu2013guide}, and providing user-intention-driven visualizations of query results \cite{ali2016big}. Thus, a comprehensive automation system needs to encapsulate intention understanding, data retrieval and analysis, and the generation of visual outputs, which traditionally requires expert involvement.

\begin{figure*}
\centering
\includegraphics[width=0.95\linewidth]{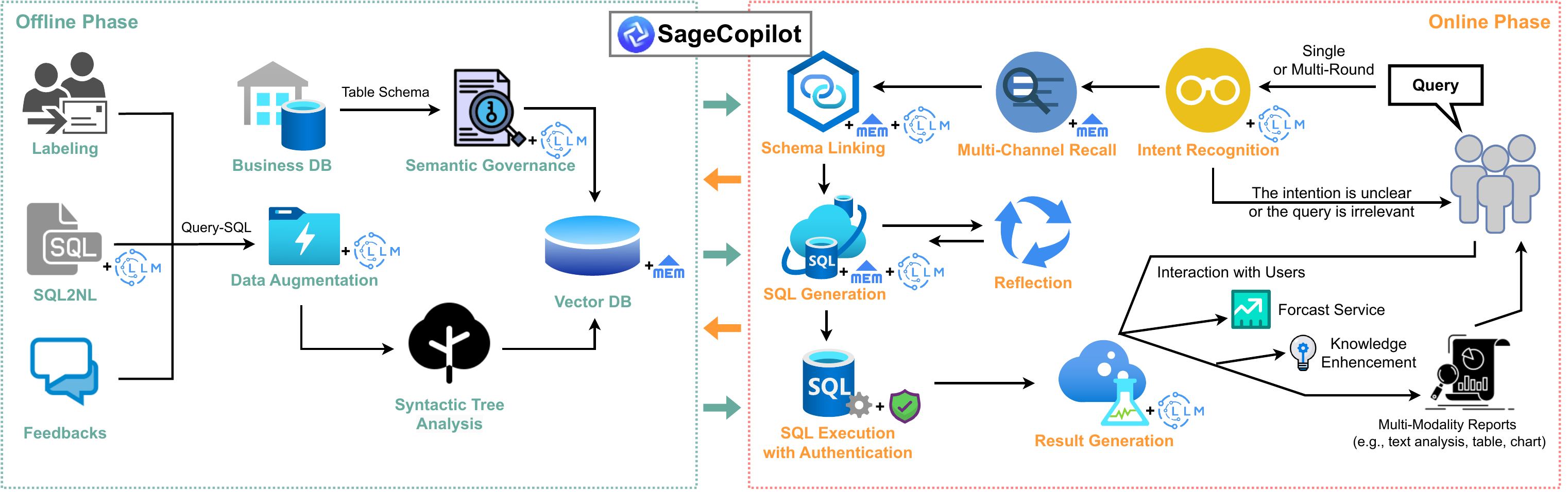}\vspace{-10pt}
\caption{The Overall framework of \ours{}.}
\label{fig:system_arch}
\vspace{-3mm}
\end{figure*}

The integration of LLMs and AutoAgents has paved the way for establishing comprehensive data science pipelines that connect language user interfaces (LUIs), databases, and data visualization tools with LLMs~\cite{xiong2023natural}. AutoAgents can clarify user intentions using conversational interfaces and then plan for data querying, analysis, and visualization by utilizing LLM capabilities. By coordinating tasks across databases, authenticators, and visualization tools, as well as employing SQL \& scripts generation~\cite{deng2022recent,achiam2023gpt,poldrack2023ai}, and data format conversions~\cite{achiam2023gpt}, AutoAgents can fulfill complex data science requirements autonomously.  
Though techniques to improve every single component, such as NL2SQL, have been intensively studied in previous works~\cite{zhongSeq2SQL2017,gualtieri2017interfaces,xuSQLNet2017,wang2019rat,wang2020dusql,deng2022recent,achiam2023gpt,poldrack2023ai}, integrating these components is still a challenging task for achieving the goal of automated data sciences in an industry settings. Several non-trivial technical issues should
be addressed as follows.
\begin{itemize}[leftmargin=*,noitemsep,nolistsep]
    \item \emph{Closed-Loop System with Multi-Tool:}  Existing systems, such as ChatGPT or Code Interpreter, either generate scripts under users' supervision or manipulate uploaded datasets for potential data analysis. Human intervention (e.g., checking the correctness of the generated scruots, copying and pasting the scripts to a local execution environment for further data manipulations) might be indispensable between every steps of interactions. However, for full automation of data science pipelines, there needs a system that processes user requests, manages databases, executes scripts, and reports analysis results in a seamless closed-loop, minimizing human intervention.
    
    \item \emph{Domain Adaptation and Generalization:}  To enhance the accuracy of script generation, supervised fine-tuning (SFT) of LLMs subject to various tasks is desired. However, SFT for adapting every data domain (e.g., databases for groceries stores, wholesales, corporate finance, etc.) or collecting a sufficiently large dataset for cross-domain generalization is resource consuming. Inappropriate SFT could hurt emergent abilities of an LLM due to catastrophic forgetting~\cite{lin2023speciality}. Thus, in addition to domain-specific SFT, the system should be capable of adapting to every data domain in the context of interactions with LLMs.
    
    \item \emph{End-to-End Correctness and Reflection:} Though every single component could be optimized with better accuracy or capacity, the system might hurt the overall user experience when it did not integrate the components well. The whole system should be aware of end-to-end correctness while some critical components should be ``self-reflectable'' towards the correctness and completeness of its input and output.
    
\end{itemize}

To achieve above goals, we propose \ours{} -- a novel, industry-grade system to automate the data science pipelines. As shown in Figure~\ref{fig:system_arch}, \ours{} consists of two phases. Given a database to be hosted by \ours{}, the offline phase first generates sufficient demonstrations for ICL to support the online phase. On the other hand,  the online phase handles the user's requests for data query and analysis while providing scripts generation, data manipulation and results visualization functionalities based on ICL with LLMs.  
The main contributions of this work are as follows.
\begin{itemize}[leftmargin=*,noitemsep,nolistsep]
    \item Given the demands of data analysis described as texts, we study the technical problem to correctly generate executable scripts that could automatically and safely run on external data tools of diverse types, such as MySQL, Spark, Hive, Flink, and eChart, within an industry-grade data science pipeline encompassing querying, analysis, visualization, authentication functionalities. To the best of our knowledge, this work is the first to incorporate LLMs, AutoAgents, LUIs, databases, visualizers, authenticators, and other tools all in a closed-loop engaged with data users, by addressing autonomy, correctness and safety issues of querying, analysis and visualization.
    
    \item To address the technical problem, we propose a novel automated data sciences system, namely \ours{}, based on newly fashioned LLM-driven AutoAgent technologies. Specifically, to maximize the accuracy of script generation while avoiding to fine-tune the LLM, \ours{} consists of two pipelines -- an offline pipeline collecting, generating and preparing a large number of demonstrations of possible user demands and the paired scripts for execution, while an online pipeline proposing LUI, NL2SQL, Text2Analyze and Text2Viz components to first identify, clarify and complete users' intentions, and then to generate the accurate scripts for querying, analysis, and visualization purposes using the prepared demonstrations through ICL. Moreover, Chain-of-Thought(COT) and prompt-tuning strategies have been adopted to improve the effectiveness of the AutoAgent through multi-round conversations between \ours{} and LLMs.
    
    \item To evaluate the performance of \ours{}, we have carried extensive experiments to test the end-to-end correctness of script generation and execution for data query, analysis and visualization, using real-world traffics. Particularly, we use ablation studies to evaluate the contribution to the end-to-end correctness made by different strategies throughout the whole data science flow. Furthermore, case studies have been done to testify the accuracy of key components, including NL2SQL, Text2Analyze and Text2Viz in Appendix \ref{sec:main_flow}. The comparisons with other prompt-based solutions, on top of the same set of LLMs, shows \ours{} achieve better end-to-end performance while maintaining good user experiences. As part of industrial contribution, we also offer our review in designing and implementing \ours{} and summarize the lessons learned.
    
    
\end{itemize}

\section{Framework Design}\label{main_method}

As depicted in Figure~\ref{fig:system_arch}, the \ours{} framework presents a sophisticated dual-phase approach to managing and processing natural language queries against complex databases. In the offline phase, significant emphasis is placed on the preparation and enrichment of a robust data foundation that underpins the online operational stage. Through metadata semantic governance, the framework lays the groundwork for a deep understanding of the data structure, ensuring comprehensive query understanding and accurate SQL generation. The construction of a rich seed data repository, supplemented by effective data augmentation strategies, further strengthens the system's ability to produce precise query responses.

Transitioning into the online phase, \ours{} seamlessly shifts from data curation to real-time user engagement. The framework's intent understanding and decision-making module utilizes the nuanced capabilities of LLM to interpret diverse user queries, cater to non-standard question phrasings, and provide contextually relevant visualizations. The system's adeptness at memory recall and schema linking further streamlines the process, enabling efficient and relevant information retrieval. The integration of SQL generation with in-context prompting, coupled with the reflective oversight offered by the SQL reflection module, ensures the production of accurate and logically coherent database queries. These features, alongside the essential components of tool use and authentication, confirm that \ours{} operates with precision, reliability, and security. Result presentation stands as the final touchpoint in the user's interaction with \ours{}, where complex data is transformed into accessible and actionable insights through text replies, visualizations, and forecast models. This not only enhances the user experience but also extends the framework's utility as a predictive tool for future trends and outcomes. The rest of this section will step into each phase for detailed design philosophies.

\subsection{Offline Phase}
\label{sec:offline}
The offline phase is devoted to the careful preparation of high-quality data that will be used in the online phase. This preparatory work involves four key activities: practicing metadata semantic governance, constructing a base of seed data, enhancing the dataset through data augmentation, and extracting information for schema linking. Once collated and refined, these datasets are embedded into a memory vector database which, in turn, equips the online phase with the necessary contextual data to meet specific user queries.

\subsubsection{Schema Semantic Governance}
Schema information is essential for data analysis tasks such as SQL writing, akin to the necessity for humans to understand table schematics. \ours{} generates complex SQL statements aligned with human intentions by interpreting schema details. A detailed exposition of table metadata is indispensable for accurately matching user intent. For fields with enumerated values, a comprehensive description of the values stored in the database and their interpretations is required. In cases involving complex data types like structures and maps, a detailed account of key attributes and their meanings is crucial. This detailed metadata aids in understanding the complex relations between attributes, allowing for accurate data retrieval in response to queries. Additionally, SQL dialects should be annotated to ensure generated SQL adheres to the specific syntax required by the target system.

\subsubsection{Seed Data Construction}
The capacity of current LLMs in generating complex SQL autonomously is fairly restricted. We leverage the ICL ability of the LLMs to bolster the stability and accuracy of SQL generation by incorporating dynamic examples of <Query, SQL> pairs relevant to the user's current question. There are three strategies for constructing these seed data pairs. The first is populating <Query, SQL> pairs common in the business's everyday workflows, especially because certain queries routinely include bespoke business indicators. Without few-shot prompting or fine-tuning mechanisms, expecting large models to autonomously generate user-specific SQL is impractical. The second method is the SQL2NL technique, which facilitates the cold start process without imposing the requirement for users to supply <Query, SQL> pairs, particularly given the abundance of SQL in the enterprise data warehouse. In this context, natural language queries are generated by LLMs based on SQL and schema information, thereby constituting <Query, SQL> pairs. The final approach involves deriving <Query, SQL> pairs from user feedback, targeting instances where LLMs have failed to produce precise results. Here, human intervention is key to correcting inaccuracies.

\subsubsection{Data Augmentation Techniques}
\ours{} incorporates two data augmentation strategies to enhance the performance and adaptability, whose ablation
study detailed in Section~\ref{sec:data_augmentation}.

\begin{itemize}[leftmargin=*,noitemsep,nolistsep]
\item Semantic-Preserving Augmentation -- the aim is to bolster the initial dataset, utilizing the LLM to transform a given query into a new version that is semantically equivalent but syntactically varied, denoted as Query’. 

\item Domain to NL\&SQL Conversion -- this strategy addresses the generation of seed datasets for new topics through a combined effort of humans and the LLMs by taking domain-specific table schemas and having the LLMs generate relevant queries. These queries are then meticulously vetted by humans to select high-quality examples that support the SQL generation process in the online phase.
\end{itemize}

\subsubsection{Schema Linking Information Extraction}
We utilize the analysis of SQL syntax trees for multiple dialects to convert our dataset from <Query, SQL> format into a <Query, Tables, Fields> format. This transformation creates a direct linkage between user-generated queries and their associated database fields.

\subsubsection{Memory Vector Database Integration}
The memory vector database is pivotal in bridging the gap between offline preparation and online functionality. It encapsulates and consolidates data into a memory structure that unites both the original and augmented data pairs—including <Query, SQL, Tables, Fields> along with <Query’, SQL, Tables, Fields>—to support robust and efficient data retrieval and management.  


\subsection{Online Phase}
\label{sec:online}

In the online phase, the system delivers a refined user experience by discerning user intent and generating precise SQL queries through a series of LLM-driven modules. The system begins with intent recognition, proceeding to link questions to database schemas, and then crafting SQL statements guided by user queries paired with similar examples. The SQL Reflection module improves accuracy by correcting errors in SQL queries. After authentication, the database runs the SQL and results are processed. The Result Generation module translates these results into textual analyses, avoiding visual complexity while still offering comprehensive insights. A Visualization component transforms data into charts. Additionally, a Forecast capability allows for the prediction of time-based data trends through simple, intuitive LLM-user interactions. Overall, the online phase is a streamlined sequence ensuring precise, user-aligned outputs, from query interpretation to actionable predictions.

\subsubsection{Intent Understanding and Decision-Making}
In authentic business environments, diverse user inquiries arise, including content beyond query parameters, topics outside the scope of the domain, incomplete queries, and follow-up questions that challenge continuity. To navigate these intricacies, an LLM-powered intent understanding and decision-making mechanism has been meticulously designed. This mechanism integrates four principal components: comprehensive intent assimilation, assessment of relevance, provision for direct query result visualization, and chart type discernment, all aligning towards heightened service precision for users.

This system harnesses user inquiries, analogous example queries, and predefined decision metrics to formulate elaborate prompts that optimize LLM responses. The comprehensive intent understanding component is calibrated to capture the essence of user requirements, addressing lacunae within multi-turn dialogues. The relevance assessment ascertains the pertinence of queries to the currently engaged topic, thereby filtering out unrelated or off-topic issues. The element of direct plotting instruction examines if an inquiry can be addressed based on preceding outputs, thereby streamlining query processing. Lastly, chart type identification tailors visualization outputs to match user requests, ensuring congruency between visual reports and user intentions.

\subsubsection{Multiple Recall \& Schema Linking}
In \ours{}, thematic domain schemas involving 1 to N tables are depicted. The challenge in practical business situations is handling expansive tables within the confined contextual bandwidth of LLM prompts. This context constraint renders the integration of comprehensive schema details for all N tables into a solitary prompt impractical. Moreover, incorporating superfluous schema information can detrimentally impact model effectiveness. As a remedy, we introduce a schema linking module preceding SQL formulation to circumvent the complications of excess schema inputs.

Contemporary schema linking interventions~\cite{wang2019rat, wang2022proton,lei2020re} often revolve around the development of fresh models, necessitating substantial resources with potential limitations in domain-agnostic generalizability. Counteracting this, our method employs a preliminary multi-recall tactic to pinpoint tables aligning with the user's query intention. The LLM is then utilized to forge links between the user's question and the relevant tables and fields.

We execute the multi-recall approach through two principal channels: (1) a similarity-based recall that aligns user queries with table schemas, and (2) a homologous recall that harnesses pre-stored schema linking data from a memory vector database initialized during an offline phase. These distinct channels function autonomously, guaranteeing that the targeted table is suitably represented within the pool of candidates. After the table recall, Schema Linking particulars are deduced from the prescribed prompts as detailed in Figure~\ref{code:schema_linking}. This dual-phase methodology fosters efficient SQL generation for voluminous tables within specific domains and streamlines the schema linking process, thereby enhancing the model's efficacy and generalization potential.

\begin{figure}[htbp]
\vspace{-10pt}
\centering
\begin{minted}[breaklines,fontsize=\scriptsize]{python}
SCHEMA_LINKING_TEMPLATE = """According to the following [schema info] and [question], return the fields related to [question] in [schema info] in ```json``` format. The ```json``` format of the returned content is as follows. It is an array, and each item must contain TABLE and FIELD. Please replace the values of variables between $$ in the following example with reasonable content:
```json
[{{"TABLE": $table_1$, "FIELD": [$field_1$, $field_2$]}}, {{"TABLE": $table_2$, "FIELD": [$field_1$, $field_2$] }}, ...]```
[schema info]:{schema_info}
{examples}
[question]: {query}
Return relevant fields:"""
\end{minted}
\vspace{-10pt}
\caption{Schema linking prompt.}\label{code:schema_linking}
\vspace{-5pt}
\end{figure}

\subsubsection{SQL Generation \& In-Context Prompting}
\label{sec:SQL_gen}
\begin{figure}[htbp]
  \centering
  \vspace{-2mm}
    \includegraphics[width=0.98\linewidth]{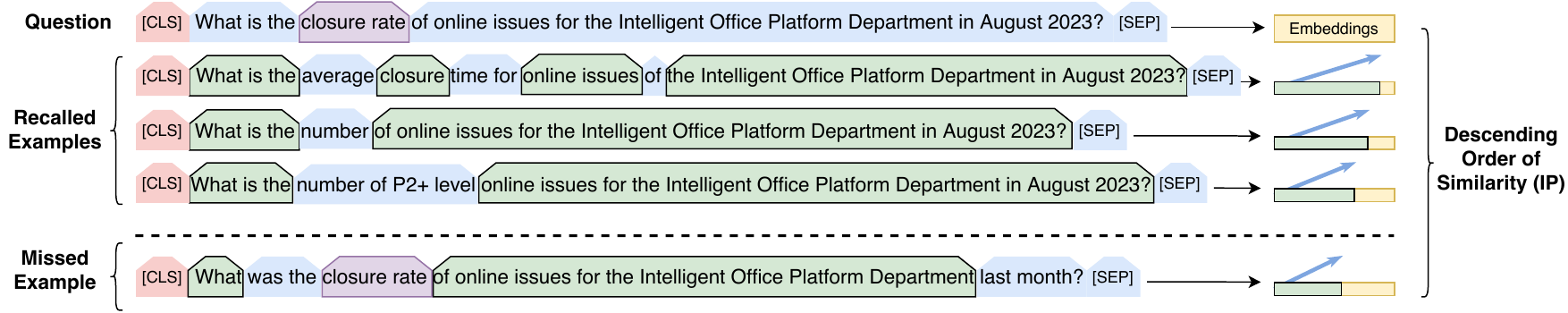}
  \vspace{-10pt}
  \caption{The miss-recalled example in similarity search.}
  \label{fig:slot}
  \vspace{-5pt}
\end{figure}

The objective of this module is to accurately generate SQL statements by fusing intent comprehension with schema linking, utilizing the in-context capabilities of LLMs. A significant technical challenge is crafting precise prompts with a restricted token input capacity. Typically, in-context examples in Figure~\ref{code:SQL_gen} are comprised of <Query, SQL> pairs retrieved from a vector database via similarity search. Initially, an input query is converted into an embedding vector\footnote{https://github.com/shibing624/text2vec}. This vector is then compared to the closest <Query, SQL> pair in the vector database by calculating the Inner Product (IP) distance, a ubiquitous measure in similarity searching. However, token input size constraints mean that it is not feasible to include all pertinent examples in the prompt template. Conventionally, only the top-K in-context examples are embedded in the template. A problematic case has been identified (as depicted in Figure~\ref{fig:slot}) where key examples may be omitted due to an unweighted IP similarity ranking, with significant terms like "closure rate" making negligible contributions to the overall similarity in comparison with other identical phrases. It has been observed that such high-ranking examples often lack the critical term "closure rate," potentially leading to incorrect SQL statement generation if the definition of "closure rate" is not present.

\begin{figure}[htbp]
\vspace{-10pt}
\centering
\begin{minted}[breaklines,fontsize=\scriptsize]{python}
SQL_TEMPLATE = """Based on the following {dialect} table and examples, generate the corresponding SQL statement for the question. Return only one SQL statement after the question, and the format must be ```sql```."""
\end{minted}
\vspace{-10pt}
\caption{SQL generation prompt.}\label{code:SQL_gen}
\vspace{-10pt}
\end{figure}

To address this problem, we have introduced a slot feature extraction method combined with usual full-length query recall. The proposed hybrid recall strategy undertakes a two-step process: (1) employ the LLM to isolate the key terms and generate a shortened query to single out the most similar example, (2) merge this kernel example based on the key term with the array of results from the full-length query recall. This method yields more robust in-context examples than the conventional approach. Additionally, the in-context functionality of the LLM is leveraged to extract patterns and interpret information from contextual question examples, thus refining the accuracy of the generated SQL statements. The superiority of this hybrid strategy is further validated in an ablation study detailed in Section~\ref{sec:slot}.

\subsubsection{SQL Reflection}
The SQL Reflection module is crafted to harness the exceptional proficiency of LLMs in amending flawed SQL expressions, significantly elevating SQL accuracy. Through detailed evaluation and comprehensive review, we scrutinize the veracity of SQL commands, delving into error etiologies encompassing verifications for table and column presence, as well as syntactic integrity. The SQL Reflection mechanism is activated solely upon the unequivocal identification of an anomaly within the SQL command.
The configuration of the prompt is meticulously engineered, delineating not just the error typologies but also encompassing the SQL queries, explicit table schemas, and in-depth diagnostic evaluations of the discrepancies. Such a framework primes the LLM to concentrate on redressing the underlying SQL inaccuracies, thus garnering an accurate and meticulously directed correction process.
The efficacy and practicability of this module have been corroborated through empirical studies detailed in Section~\ref{sec:sql_reflection}. The data derived from these experiments solidifies the premise that this modality is not merely efficacious but also eminently viable.

\subsubsection{Tool Use \& Authentication}
Our system, \ours{}, accommodates SQL command execution across a spectrum of dialects, spanning MySQL, PostgreSQL, Flink SQL, Hive SQL and Spark SQL. It is noteworthy that queries executed via MySQL and PostgreSQL can deliver output at the precision of milliseconds. Nevertheless, Spark/Hive/Flink SQL execution duration may range from a few minutes to hours, influenced by variables including data scale and the intricacy of the SQL query. An integral component of \ours{} is its granular authentication mechanism, aligned with the zero-trust security framework. Prior to the execution of any SQL query, the system rigorously verifies whether users hold the necessary authorizations for access to the tables and columns referenced in the SQL statement within the allocated milieu. Progression to the execution phase is predicated on the assertive validation of such privileges. Conversely, if the user is found deficient in the required permissions, \ours{} immediately communicates the execution impediment to the user, citing inadequate authorization. This vigilant and precise authentication protocol is in strict adherence to the tenets of the zero-trust model, bolstering security protocols and assuring that SQL execution transpires strictly within the ambit of sanctioned parameters.  

\subsubsection{Result Generation}
\label{subsec:result_generation}
Upon the retrieval of execution outcomes from the database engine, the result generation module commences its role in constructing the conclusive output for web-based applications. Demonstration of this process is evident in the lower right segment of Figure~\ref{fig:system_arch}, wherein a diversely formatted report analogous to a portfolio is composed. Such a report encompasses textual analysis, visual representations through graphs and charts, and tabular data. Additionally, this phase may initiate further user interaction through the integration of predictive services and knowledge-driven analytical processes.

\paragraph{Text Analysis}
\label{para:text_analysis}
Predominantly, the result generation module acting online employs a LLM agent for the transformation of execution results, articulated in markdown or JSON-esque structures, into textual discourse. The composition of the textual output is not limited to the portrayal of SQL execution outcomes but also encapsulates a synthesized overview and preliminary analysis of the data. The prompt utilized for guiding the LLM agent is formulated in Appendix~\ref{sec: text_analysis}. In such an approach, the aspect of visualization has been deliberately decoupled, as a distinct module with an exclusive focus on graphical representation has been specifically crafted to augment the inherent capabilities of chart formation to their fullest extent without complication. It warrants attention that under certain circumstances, users may venture inquiries pertaining to key, yet uncomputed, metrics—take for instance the "closure rate", potentially non-existent within the current table schema. This presents a challenge as the LLM may find itself incapable of delivering the anticipated response. A remedy to this quandary is the activation of a knowledge enhancement mechanism, prompting the user for the precise formula necessitated for the computation of the "closure rate". Upon acquisition of the accurate or anticipated formula, the LLM is equipped to amalgamate it with the SQL execution findings, thereby generating an apt and tailored response.

\paragraph{Visualization}
In order to refine front-end processing and deliver a more comprehensible presentation of graphical data, we utilize the capabilities of an LLM agent to generate bar, line, and pie chart code in Echarts-compatible JSON format. However, due to the broad range of outputs possible from the LLM agent, the produced visualizations may not consistently meet the user's precise needs. To address this, we leverage the ICL of the LLM to adapt the prompt with examples that anticipate the user's preferred chart types. This method is instrumental in enhancing the standardization and precision of the LLM-generated outputs, thus better fulfilling user expectations. One such prompt is detailed in Appendix~\ref{sec: chart_generation}.

\paragraph{Forecast}
\ours{} is equipped to conduct analyses and forecasts based on time-related data retrieved via SQL. We have explored various lightweight models suitable for temporal data analysis and forecasting, including Prophet. Leveraging the LLM's ability to interpret user's natural language and convert it into API calls for time series analysis models simplifies complex functions such as trend and periodicity estimation, as well as forecasting future events across different timescales. Due to constraints on length, we illustrate only a single real-world online case study of forecast using Prophet in Appendix~\ref{fig:prediction_prophet}.  

\section{deployment and evaluation}\label{evaluate}
In the following section, we present the intricate details of deploying our novel \ours{} within an industrial environment and carry out extensive experimental evaluations to assess the system's overall performance and the impact of each optimization technique.

\begin{figure}[htbp]
    \vspace{-5pt}
    \centering
    \includegraphics[width=0.85\linewidth]{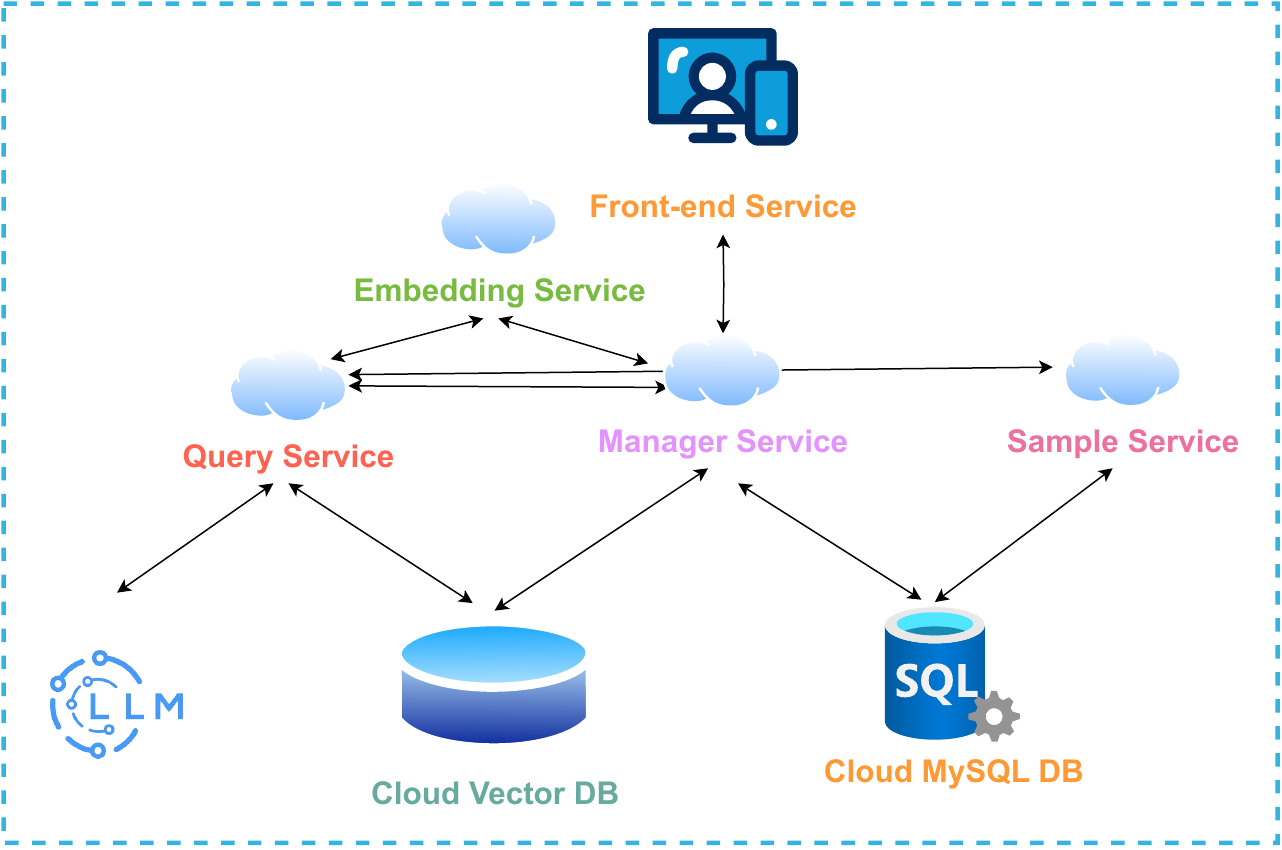}
    \vspace{-10pt}
    \caption{System deployment architecture}
    \label{fig:system-deployment-architecture}
    \vspace{-10pt}
\end{figure}

\noindent\textbf{Setups.} 
Figure \ref{fig:system-deployment-architecture} depicts the deployment of \ours{} using a microservice architecture within a Kubernetes (K8s) cluster, with service configurations detailed in Table \ref{tab:system_services_description} in Appendix~\ref{sec:sys_config}. To enhance the system's stability and performance, \ours{} strategically avoids direct connections between Python web services, such as the query service, and the MySQL database. Instead, a Java-based web service, namely the manager service, acts as an intermediary, managing front-end user requests and SQL database interactions. This design choice aims to enhance system throughput and reduce stability risks associated with direct Python-MySQL connections. Additionally, a MySQL database is implemented prior to employing operations within the vector database to maintain index uniqueness. Based on this foundational deployment, we perform a series of experiments to ascertain the efficiency and efficacy of the deployed system and to conduct an ablation study for each optimization strategy in \ours{}.


\noindent\textbf{Datasets.} 
The \textbf{\DuSQL{}} dataset, frequently used to evaluate the accuracy of Text-to-SQL parsing across various databases, challenges models to adapt to new database schemas. The dataset contains 23,797 high-quality Chinese Text-SQL pairs over 200 distinct databases. In this investigation, we opted for 10 databases from DuSQL to constitute the thematic domain in \ours{}, following the offline phase described in Section~\ref{sec:offline}. The dataset was segmented into 700 exemplar set entries and 215 test set entries.

We developed the \textbf{\industrial{}} from the data management platform of Baidu Inc. Owing to privacy concerns, the obtained data primarily consisted of redacted real tables. Utilizing these sanitized tables and through manual curation, we composed 6 thematic domains that include 10 data tables. The dataset was segmented into 212 exemplar set entries and 61 test set entries, with each entry comprising both the input queries and the corresponding SQL query statements.

\begin{figure}[htbp]
\vspace{-10pt}
\centering
\subfigure[\industrial{}]{
\includegraphics[width=0.22\textwidth]{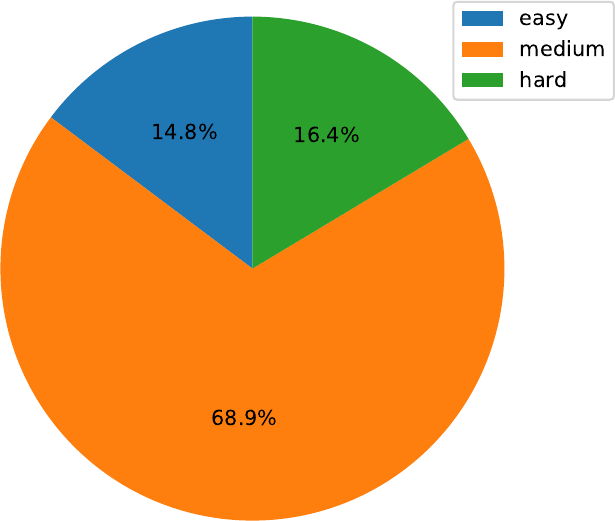}
}
\subfigure[\DuSQL{}]{
\includegraphics[width=0.22\textwidth]{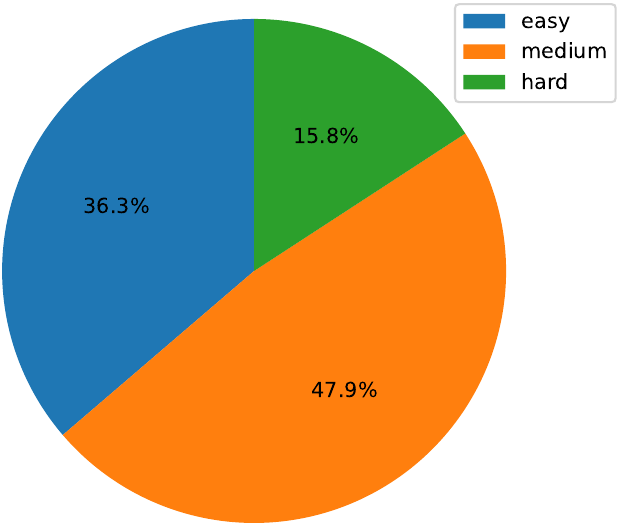}
}
\vspace{-15pt}
\caption{The difficulty of \industrial{} and \DuSQL{}.}
\label{fig:difficulty}
\vspace{-10pt}
\end{figure}

To evaluate the difficulty of the \industrial{} and \DuSQL{} datasets, we employed the NL2SQL evaluation methodology provided by the BIRD benchmark~\cite{li2023llm}, which considers four dimensions: comprehension of the question, the requirement for external knowledge, complexity of the data, and complexity of SQL. Each dimension was rated on a scale from 1 to 3. Samples accumulating a total score under 4 were deemed easy, those between 5 and 6 were regarded as of medium difficulty, and samples with scores of 7 or higher were classified as challenging. Figures \ref{fig:difficulty} illustrate the difficulty distribution across the datasets.

\medskip
\noindent\textbf{Metrics.}
To assess the performance of the proposed \ours{}, we evaluate it from two perspectives: (1) the quality of SQLs generated in the \ours{} pipeline, and (2) the quality of answers as an automated data analysis tool. For the first aspect, we adopt the mainstream NL2SQL metrics such as exact match, execution accuracy. We also explore one human-aligned metric as one supplementary indicator. For the second one, we cooperate a team of experts (trained with the domain knowledge) to rate the answers based on a panel of human-designed criteria. The detailed definition for each metric is illustrated as follows:

\begin{itemize}[leftmargin=*,noitemsep,nolistsep]
    \item \textbf{Exact Match (EM)}, namely the percentage of questions whose predicted SQL query is equivalent to the gold SQL query, is widely used in text-to-SQL tasks. Suppose that $Y_i$ is the i-th ground truth SQL and $\hat{Y_i}$ is the i-th predicted SQL, EM can be computed by $EM=[\sum_{i=1}^{N}\vmathbb{1}(Y_i,\hat{Y_i})]/N$.
    \item \textbf{Execution Accuracy (EX)}, namely the percentage of questions whose predicted SQL obtains the correct result, assumes that each SQL has an result. Considering the result $V_i$ is executed by the $Y_i$ and $\hat{V_i}$ is executed by the $\hat{Y_i}$, EX can be computed by $EX = [\sum_{i=1}^{N}\vmathbb{1}(V_i,\hat{V_i})]/N$, where $\vmathbb{1}(\cdot)$ is an indicator function, which can be represented as $\vmathbb{1}(V,\hat{V})=\left\{\begin{matrix} 1, V=\hat{V}\\ 0, V\neq{\hat{V}}\end{matrix}\right.$.
    \item \textbf{Human-aligned Accuracy (HA)} is introduced to reconcile the discrepancies between traditional benchmark evaluation metrics and actual human preferences. To illustrate in Figure~\ref{code:badcase_ex}, consider a case from the test set:
    \begin{figure}[htbp]
    \vspace{-10pt}
    \centering
    \begin{minted}[breaklines,fontsize=\scriptsize]{SQL}
    -- Question: Which grade has the most high schoolers?
    -- pred:
    SELECT grade, COUNT(*) AS num_highschoolers FROM Highschooler GROUP BY grade
    ORDER BY num_highschoolers DESC LIMIT 1;
    -- gold:
    SELECT grade FROM Highschooler GROUP BY grade ORDER BY count(*) DESC LIMIT 1;
    \end{minted}
    \vspace{-10pt}
    \caption{Badcase for EX metric.}\label{code:badcase_ex}
    \vspace{-10pt}
    \end{figure}
    
    In this instance, the query produced by the LLM (listed first) was identified as erroneous based on the established gold standard (listed second). Notably, the generated query includes an additional computed column indicating the count of students, presenting a more descriptive outcome. According to traditional Evaluation Accuracy (EX), such an informative response would be deemed incorrect. To address and potentially ameliorate this sort of misalignment, HA has been devised as a means to measure the extent to which generated SQL queries align with human-alike judgment. Specifically, HA evaluates the suitability of a generated SQL query by executing both the forecasted and gold standard queries, and using their respective outputs to generate natural language responses to the input question. The accuracy of a generated query is then assessed by determining whether the natural language answer derived from the output of the predicted query is correct, taking into account the input question and the reference answer produced from the output of the gold standard query. The response acquired from the language model is binary: Yes or No.
    \item \textbf{Artificial Assessment (AA)} is a manual expert scoring index, composed of evaluations across multiple dimensions, including two text response indicators and three visual chart indicators. The text response indicators consist of data consistency and response richness, while the visual chart indicators consist of data consistency, display capability, and display rationality. A detailed description can be found in the scoring Table~\ref{tab:AA_metric} in the appendix. We will also calculate the total score to demonstrate comprehensive performance assessment.

\end{itemize}

\noindent\textbf{LLM Models.}
In the evaluation of fundamental NL2SQL task performance, ErnieBot's capabilities were compared with other accessible open-source models, with focus on features such as fine-tuning and in-context learning proficiencies. As indicated in Table~\ref{tab:raw_llm_baselines}, ErnieBot demonstrates superior execution accuracy (EX), particularly in Few-Shot Prompting scenarios, affirming its robustness in this context. It is important to note that the ErnieBot series serves as the foundational Large Language Models (LLMs) for our system, \ours{}. Furthermore, \ours{} is designed to be adaptable, capable of augmenting the inherent functionalities of various other LLMs as alternatives, thereby extending the range of their native abilities.

\begin{table}[h]
    \vspace{-5pt}
    \caption{LLM base execution accuracy (EX) comparison.}
    \vspace{-10pt}
    \centering
    \scalebox{0.65}{
    \begin{tabular}{cccccc}
        \hline
        & Erniebot & ChatGLM-6B & ChatGLM-6B & ChatGLM2-6B & Llama-Chinese-Alpaca \\
        & Raw & Raw & Finetuned & Raw & Finetuned\\
        \hline
         Zero-Shot  & 25.0\% & 0.0\% & 19.0\% & 0.0\% & 12.0\%  \\
         \hline
         Few-Shot   & 69.0\% & 44.0\% & 50.0\% & 37.5\% & 56.0\%  \\
        \hline
    \end{tabular}}
    \label{tab:raw_llm_baselines}
    \vspace{-10pt}
\end{table}

\subsection{Overall Performance}
In Table~\ref{tab:overall}, we present the experimental results of the system's overall online serving performance when leveraging the framework designated as \ours{}. The assessed components of the experiment revolve around SQL query accuracy and the multifaceted Artificial Assessment metric, which incorporates evaluations from both text and visual perspectives.

Key findings indicate that \ours{} achieved moderate SQL Accuracy, with Exact Match (EM) at 45.6\%, Execution Accuracy (EX) at 72.6\%, highlighting correct results despite non-exact matches, and Human-aligned Accuracy (HA) slightly higher at 77.8\%, showing competency in complex queries. The Artificial Assessment yielded a high score in Text Data Consistency (T1) at 3.26/4, but lower in Text Richness (T2) at 1.42/3 since we specify the briefness in the prompt~\ref{code:text_analysis}. Visual Charts evaluation gave middling scores: Data Consistency (V1) at 1.34/2, Display Capability (V2) at 0.73/1, and Display Rationality (V3) at 1.24/2, suggesting reliable numerical data representation but areas to enhance in visual display and chart selection. The cumulative score for the Artificial Assessment(AA), denoted as SUM, adds these individual aspects to form an aggregate score. For \ours{}, a SUM of 7.95 was obtained, showcasing the system's notable but imperfect capability in delivering both textually and visually coherent data representations. Considering multi-round querying as a more complex task than single-round querying due to the iterative nature of refining queries based on user feedback and maintaining context over multiple interactions, the performance measures reflected in AA demonstrate the challenging nature of multi-round querying.

\begin{table}[htp]
    \vspace{-5pt}
    \centering
    \caption{\small{Overall performance of online serving with \ours{}.}}
    \vspace{-10pt}
    \scalebox{0.72}{
    \begin{tabular}{cccccccccc}
        \hline
         \multirow{2}[3]{*}{Dimension} & \multicolumn{3}{c}{SQL Accuracy} & \multicolumn{6}{c}{Artificial Assessment (avg. score)} \\
         \cmidrule(l){2-4}  \cmidrule(l){5-10} 
         & EM & EX & HA & T1 & T2 & V1 & V2 & V3 & SUM \\
         \hline
         \industrial{} & 45.6\% & 72.6\% & 77.8\% & 3.26 & 1.42 & 1.34 & 0.73 & 1.24 & 7.95 \\
         \hline
         multi-round & - & - & - & 2.17 & 1.18 & 0.85 & 0.45 & 0.81 & 5.48\\
         \hline
        \end{tabular}}
    \label{tab:overall}
    \vspace{-10pt}
\end{table}



\subsection{Ablation Test}
\subsubsection{Data augmentation}
\label{sec:data_augmentation}

In the conducted ablation studies, we sought to evaluate the efficacy of our method, denominated as \ours{}, through different data augmentation strategies. The experiments were categorized into four distinct sets: i) the zero-shot approach, devoid of any exemplar collection, relying solely on the language model's inherent capabilities; ii) the ER method, which leverages exemplar sets to invoke the model's in-context learning by recalling similar instances; iii) the ER+SA method—our proposed approach—where similar instances are retrieved from a semantically augmented exemplar set; iv) the ER+D2N approach, where the recall procedure is applied to an exemplar set enriched with domain-specific NL2SQL augmentations.

The outcomes, detailed in Table \ref{tab:4_1}, indicate a marked enhancement in both EM and EX when the ER process is employed. Incremental improvements were further observed with our proposed ER+SA strategy, albeit to a lesser extent. However, the introduction of the domain to NL\&SQL data augmentation (ER+D2N) did not culminate in any notable performance gains. The stagnation observed with the ER+D2N augmentation suggests that the similarity in distribution between the augmented and test samples was too great to provide a significant benefit. In contrast, the implementation of both example recall and semantic-preserving augmentation—the cornerstone techniques of \ours{}—demonstrate effective enhancements to the model's performance.  

\begin{table}[]
\vspace{-5pt}
\caption{The comparison of data augmentation strategies.}
\label{tab:4_1}
\vspace{-10pt}
\scalebox{0.72}{
\begin{tabular}{lllllllll}
\hline
\multirow{2}{*}{dataset} & \multicolumn{2}{c}{Zero-shot} & \multicolumn{2}{c}{ER} & \multicolumn{2}{c}{ER+SA} & \multicolumn{2}{c}{ER+D2N}\\ \cline{2-9} 
                         & EM           & EX             & EM           & EX           & EM           & EX        & EM          & EX  \\ \hline
Industrial               & 0\%          & 20.9\%         & 45.6\%       & 72.6\%       & 51.6\%       & 73.0\%    & 45.6\%      & 72.6\%  \\
\hline
DuSQL                    & 0\%          & 0\%            & 11.5\%       & 45.9\%       & 16.4\%       & 60.7\%    & 9.8\%       & 45.9\%  \\ \hline
\end{tabular}
}
\vspace{-10pt}
\end{table}

\subsubsection{SQL2NL}
\label{sec:sql2nl}

The SQL2NL mechanism substantively augments the automation of data augmentation and the optimization of tasks post-system deployment. Implementing SQL2NL-enhanced examples improves the Execution (EX) score for \ours{} from 56.2\% to 81.3\% within the chain supermarket domain, part of a broader \industrial{}, where the accuracy of SQL2NL achieves 92.0\%. To validate SQL2NL's efficacy, the recall capability of the vector database was tested by mixing SQL2NL-generated examples with irrelevant ones within the domain. Starting with 16 accurate and executable SQL queries, a Language Model produced three related questions per query, yielding 48 SQL2NL-enhanced questions. These were introduced into vector databases with incremental additions of unrelated examples, as shown in Figure~\ref{fig:sql2nl}. The data bolstered by SQL2NL exhibited high congruity in aligning with the original questions, even amidst a burgeoning presence of extraneous data, thereby validating SQL2NL's efficacy as a potent mechanism for generating foundational seed data. The LLM prompt configuration is available in Appendix~\ref{sec: sql2nl}.


\subsubsection{Schema Linking.}
\label{sec:schema_exp}
\begin{figure}[htbp]
\vspace{-10pt}
\centering
\begin{minipage}{.23\textwidth}
  \centering
  \includegraphics[width=.98\linewidth]{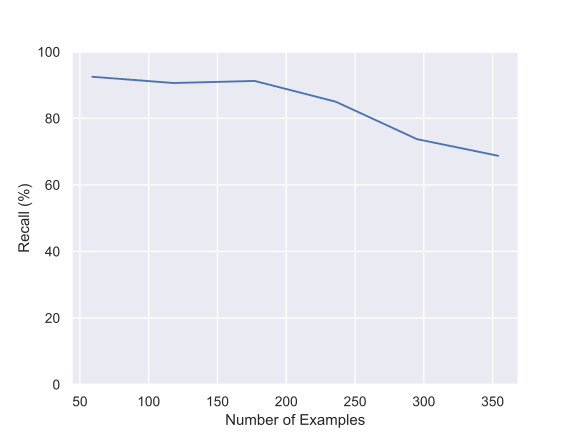}
  \vspace{-10pt}
  \captionof{figure}{The recall rate of examples using SQL2NL.}
  \label{fig:sql2nl}
\end{minipage}%
\hspace{8pt}
\begin{minipage}{.22\textwidth}
  \centering
  \includegraphics[width=.88\linewidth]{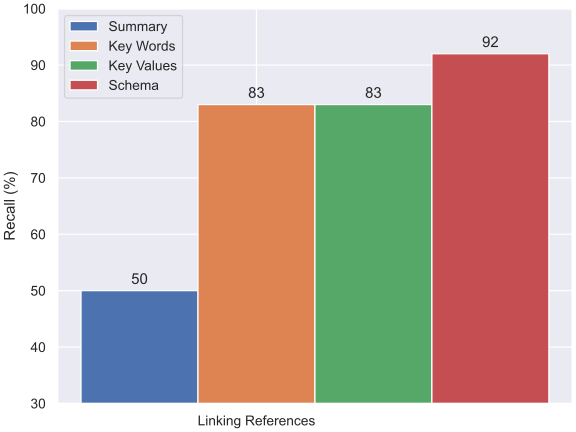}
  \vspace{-10pt}
  \captionof{figure}{Comparison of schema linking Strategies.}
  \label{fig:schema_exp}
\end{minipage}
\vspace{-10pt}
\end{figure}

In the context of multiple recall, each retrieval channel should strive to maintain independence to ensure that the target table can appear in the candidate set. The evaluation mainly focuses on two aspects: (1) whether the target table appears in the candidate set; and (2) the desirability of the target table appearing towards the beginning of the candidate set. Figure~\ref{fig:schema_exp} presents the results of four retrieval experiments conducted on 100 actual business tables, with most tables containing over 100 fields. The experimental results indicate that the retrieval strategies achieved recall of 50\%, 83\%, 83\%, and 92\% respectively. This suggests that the strategy based on direct table schema similarity performed notably better than others based on summarization of schema details, key words and the values of the key words. It's also important to consider the potential implications of the experimental results. The high recall of the direct table schema similarity strategy indicates its effectiveness in identifying relevant tables and columns for a given query. In contrast, the lower recall achieved by the summarization-based strategies may suggest limitations in capturing the nuanced details necessary for precise retrieval.

\subsubsection{Slot Feature Extraction.}
\label{sec:slot}
In response to the observed challenge where a user's query regarding "close ratio" is obscured by extraneous language (e.g., in Appendix~\ref{case:key_word}), resulting in potential retrieval of non-germane instances by schema matching techniques, we introduce the Slot Feature Extraction (SFE) approach. This method's efficacy is rigorously evaluated through a direct comparison, utilizing EM and EX metrics, on a targeted subsection of our comprehensive real-traffic dataset. The selected subset comprises 81 samples belonging to a domain noted for its susceptibility due to the relatively brief nature of the key terms within broader textual contexts. Our empirical study, discussed in Section~\ref{sec:SQL_gen}, reveals that when the Inner Product (IP) similarity metric is employed for example retrieval, there is a heightened probability of mismatches particularly when the key phrase is less prevalent within the data. Such circumstances are graphically depicted in Figure \ref{fig:slot}, highlighting how these mismatches can precipitate a significant decline in the recall rate of relevant in-context examples from this particular dataset sector.  

The implementation of our method without SFE (SFE w/o) served as the baseline for the comparison. This baseline delivered an EX of 53.1\% and an EM of 81.5\%, revealing the inherent challenge in accurately matching short key phrases within the data. Upon integrating the SFE technique (SFE w/), we observed a notable improvement in performance, where EX improved by 7.0 percentage points, reaching 56.8\%, and EM increased by 4.5 percentage points, achieving an 85.2\% success rate. These enhancements underscore the significance of the SFE approach in boosting the system's capability to discern and correctly retrieve pertinent instances based on key terms.



\begin{table}[htbp]
\vspace{-5pt}
\centering
\begin{minipage}{.21\textwidth}
  \captionof{table}{Ablation study with slot feature extraction (SFE) strategy.}
  \vspace{-10pt}
  \centering
  \scalebox{0.7}{
    \begin{tabular}{ccc}
        \hline
        Plan & EM & EM \\
        \hline
         SFE (w/o)  & 53.1\% & 81.5\%    \\
         \hline
         SFE (w/)   & 56.8\% & 85.2\%   \\
         \hline
        DIFF    &  7.0\%$\uparrow$ & 4.5\%$\uparrow$   \\
        \hline
    \end{tabular}}
  \label{tab:slot_feature_ab}
\end{minipage}%
\hspace{5pt}
\begin{minipage}{.21\textwidth}
\captionof{table}{Ablation study with SQL reflection strategy.}
\vspace{-10pt}
  \centering
  \scalebox{0.7}{
    \begin{tabular}{ccc}
        \hline
        Plan & EM & EX \\
        \hline
         SQL Reflection (w/o)  & 15.0\% & 65.0\%    \\
         \hline
         SQL Reflection (w/)   & 10.0\% & 80.0\%   \\
         \hline
        DIFF    &  5.0\%$\downarrow$ & 15.0\%$\uparrow$   \\
        \hline
    \end{tabular}}
  \label{tab:sql_reflection_ab}
\end{minipage}
\vspace{-10pt}
\end{table}


\subsubsection{Reflection on SQL Generation}
\label{sec:sql_reflection}

In our study, we conducted an ablation experiment to validate our SQL Reflection methodology by addressing diverse SQL errors such as syntax inconsistencies, references to non-existent tables, and omissions of requisite columns. Our dataset, composed of 20 distinct samples, employed Evaluation Metrics (EM) and Execution Accuracy (EX) as benchmarks for assessing our strategy's efficacy. The SQL Reflection strategy, which operates without detailed SQL error annotations, was employed as the baseline under review and is summarized in Table~\ref{tab:sql_reflection_ab}. This baseline registered an EM score of 65\% and an EX rate of 15\%. However, upon the integration of explicit SQL error descriptions into the learning model, our records indicated a modest decline in EX by 5\%, whilst EM exhibited a notable improvement of 15\%. These findings substantiate the effectiveness of descriptive error integration in SQL query generation methodologies.  

\subsubsection{Effect of Miscellaneous Designs}
The system enhances the performance of LLMs through two innovative forms of feedback: human-to-machine and machine-to-human. By correcting errors in SQL responses manually and feeding these corrections back into the system, human-to-machine feedback significantly improves system accuracy, as visually represented in Figure~\ref{fig:human_to_machine_feedback}. Meanwhile, machine-to-human feedback recognizes the need for human intervention when the system's responses may not be accurate. In such cases, the dialogue system seeks additional information from a human to clarify ambiguous queries. This ensures responses are precise and tailored to the user's intended meaning, creating a dynamic, iterative loop that continuously refines the system's performance and user experience. We show the supplementary results in Appendix~\ref{sec:feedbacks}.

\subsection{Lessons Learned and Discussions}

In this section, we outlined the deployment and experimental evaluation of our advanced system, \ours{}, in an industrial setting. Our exploration covered the system architecture, datasets, performance metrics, and the impact of design enhancements. Results showed that while not every SQL query was an exact match (EM), correct results (EX) were frequently produced, indicating that the system can tolerate some structural variances. The overall assessment confirmed the system's consistency in generating text responses and charts, though it suggested that analysis depth and chart selection could be improved. Ablation studies revealed that methods like example recall and schema similarity significantly affected system performance, with SQL2NL and Slot Feature Extraction improving EX scores by aiding in accuracy. Evaluating SQL generation strategies also demonstrated the benefits of incorporating error descriptions for better EM rates. Overall, the evaluation highlighted effectiveness of \ours{} in SQL generation and data analysis, as well as the significance of human feedback mechanisms in refining precision, thereby providing insights into enhancing LLM systems in complex industrial environments.

In response to the limitations in LLM capabilities and the complexity of SQL data within the industrial sector, we have developed an array of engineering solutions and learned lessons from realistic deployment. \textbf{Lesson 1:} Confronting the issue of restricted token input length for LLM (that make it impossible to put all schemes and attribute descriptions into one prompt), our research has adopts an innovative ``multiple recall and schema linking'' paradigm. This technique facilitates the association of user inquiries to expansive datasets, including numerous tables and columns. \textbf{Lesson 2:} In scenarios where SQL queries are characterized by complexity and include shared components, an advanced strategy was employed: the restructuring of original SQL queries through the creation of views. This strategic alteration is cataloged in Appendix~\ref{case:sql_view} and has resulted in a significant enhancement of end-to-end query accuracy, approximately by 50\% in our experiments. \textbf{Lesson 3:} The end-to-end correctness requires not only self-reflection, but also the continuous collection of demonstrations for ICL, as feedback from the online phase, to improve the system accuracy.




\section{Conclusion}
In summary, this paper introduces \ours{}, an advanced, industry-grade system that offers automated data science pipeline by effectively integrating LUIs, AutoAgents, databases, data visualizers and LLMs. It delivers an end-to-end solution capable of handling natural language instructions on querying, analysis, and visualization tasks with minimal human intervention. \ours{} incorporates a two-phase design -- an online component refining users' inputs into executable scripts through In-Context Learning (ICL) and running the scripts for results reporting \& visualization, and an offline preparing demonstrations requested by ICL in the online phase. Various prompt-tuning strategies have been used to augment \ours{} for enhanced performance. Our rigorous evaluation across real-world scenarios demonstrates the advantages of \ours{} in generating/executing scripts and offering results with visualization. Open issues and the lesson learned have been discussed as part of our contribution from industry perspectives.

\bibliography{Article}
\bibliographystyle{ACM-Reference-Format}

\appendix

\begin{table*}
    \centering
    \caption{\small{Overview of the system deployment}}
    \vspace{-10pt}
    \scalebox{0.70}{
    \begin{tabular}{cp{4cm}p{7cm}p{3cm}p{4cm}p{2cm}}
        \hline
        Service & Front-end & Manager & Sample & Query & Embedding \\
        \hline
        Description & Responsible for the front-end page requests. & Task scheduling management to cope with LLM API QPS limitations; Theme domain management. & Provide a routine data augmentation service. & Interactions with LLM and vector db based on the LangChain. & Turn the Query into a vector. \\
        \hline
        Number of Pods & 1 & 2 & 1 & 2 & 1 \\
        \hline
        CPU Cores(vCPU) & 3 & 4 & 4 & 8 & 6 \\
        \hline
        Memory(Gi) & 12 & 8 & 8 & 16 & 16 \\
        \hline
        Storage(Gi) & 70 & 70 & 70 & 170 & 170 \\
        \hline
        Language & JavaScript & Java & Java & Python & Python \\
        \hline
        Framework & React & Spring Boot & Spring Boot & FastAPI & FastAPI \\
        \hline
    \end{tabular}}
    \label{tab:system_services_description}
\end{table*}

\begin{table*}
    \centering
    \caption{\small{Scoring system for the AA metrics.}}
    \vspace{-10pt}
    \scalebox{0.70}{
    \begin{tabular}{cp{5.3cm}p{5.2cm}p{3cm}p{3cm}p{5cm}}
        \hline
         \multirow{2}[3]{*}{Dimension} & \multicolumn{2}{c}{Text Reply (T)} & \multicolumn{3}{c}{Visual Chart (V)} \\
         \cmidrule(l){2-3}  \cmidrule(l){4-6} 
         & Data Consistency (T1) & Richness (T2) & Data Consistency (V1) & Display Capability (V2) & Display Rationality (V3) \\
         \midrule
         Description & Consistency and rationality of numerical data within the response. & Whether the text reply include additional information beyond a direct statement, e.g., analysis, summary &
         Consistency of numerical data between the table and the chart. & Whether the chart is displayable. & 
         Whether the chart's type is the best representation for the current question. \\
         \midrule
         Criteria & \textbf{Scoring (0-4):} & \textbf{Scoring (0-3):} & \textbf{Scoring (0-2):} & \textbf{Scoring (0-1):} & \textbf{Scoring (0-2):}\\  
         &0: Completely incorrect data & 0: No additional information & 0: Inconsistent & 0: Not displayable & 0: Irrational (e.g., trends as pie charts)\\
         &1: Incorrect data or irrelevant numbers & 1: Incorrect additional information & 1: Partially consistent & 1: Displayable & 1: Moderate (e.g., trends as bar charts)\\
         &2: Correct but includes irrelevant numbers & 2: Accurate but simplistic additional info & 2: Fully consistent & & 2: Rational (e.g., trends as line charts)\\ 
         &3: Correct data, irrational expression & 3: Diverse and accurate additional info\\ 
         &4: Both data and expression are rational & \\
         \hline
        \end{tabular}}
    \label{tab:AA_metric}
\end{table*}

\section{Related Work}\label{related_work}

The quest to bridge the gap between natural language and database querying has led to the emergence of the NL2SQL field, which seeks to translate natural language questions into executable SQL queries. The pioneering works of Zhong et al. \cite{zhongSeq2SQL2017} with Seq2SQL and Xu et al. \cite{xuSQLNet2017} with SQLNet laid the foundation for subsequent advancements, integrating reinforcement learning and sequence-to-sequence models to adapt to the nuanced complexity of natural language. The integration of language models as agents, such as through LangChain, underscores a next-generation approach to human-computer interaction. Modern implementations, as described by Shuster et al. \cite{shuster2021langchain}, showcase the efficiency of these agents in not only understanding but also executing language-based tasks within software environments. Large language models (LLMs), including GPT variants, have demonstrated considerable effectiveness in NL2SQL applications through fine-tuning. Papers like that of Wang et al. \cite{wang2020tuninggpt} detail the intricate prompt engineering required to guide these models toward generating syntactically and semantically accurate SQL queries. The importance of prompt engineering within the human-machine collaboration framework cannot be overstated. As investigated by Webb et al. \cite{webb2020prompt}, well-engineered prompts drastically enhance a model's performance, acting as a form of 'soft' programming that specifies the task at hand without the need for hard-coded algorithms.

Furthermore, the field of Natural Language Interfaces for Data Science promotes an accessible environment for data analysis, as highlighted by Gualtieri's work \cite{gualtieri2017interfaces}. Such interfaces are increasingly sought after for their ability to democratize data science, enabling users with limited technical background to harness complex data analytics through conversational engagement. In the compilation of this literature, it is evident that the maturation of NL2SQL technologies is closely entwined with the evolution of LLMs, agent frameworks, and methodologies in prompt engineering. Amidst this synergy, the overarching goal remains to simplify and streamline the end-user experience in interacting with databases and performing data science tasks.


\section{System Configuration \small{[Table~\ref{tab:system_services_description}]}}
\label{sec:sys_config}

\section{Main Flow} 
\label{sec:main_flow}
\subsection{Intent Understanding and Decision-Making} In practical business settings, users often engage in multi-turn dialogues, posing follow-up questions subsequent to an initial inquiry. These successive questions, however, are frequently characterized by missing context or complete departure from the original topic. Mishandling such queries as definitive intents can result in erroneous outcomes and needless processing time. For instance, following an initial question about authorization request counts in October '22, a user may ask, "What about June?" or prompt for visual representations like "Could you draw a bar chart?" or pose unrelated questions such as "How's the weather today?". To navigate these multifaceted interactions, a systematic approach to intent understanding and decision-making is crucial. This involves methods for comprehensive interpretation to supplement incomplete follow-ups, the ability to directly generate visualizations from query results, assistance in chart type selection for optimal data portrayal, and relevance filtering to appropriately address or dismiss off-topic queries.


\subsection{SQL Generation}

Delving into the industry realm, consider a user question such as "What is the trend in employee [Zhou Hui]'s monthly sales in [the second half of 2022]?" Initially, we extract the pertinent schema information from the domain of the user's query, analogous to the procedures for a chain supermarket domain. We then compute the similarity between the user's question and data vectors in the vector database to select the most suitable examples containing <Query, SQL> pairs, which is shown in Figure~\ref{code:examples}. These examples serve as contextual aid in formulating a precise SQL statement. 

\begin{figure}[htbp]
\vspace{-10pt}
\centering
\begin{minted}[breaklines,fontsize=\scriptsize]{SQL}
--Query1: What is the trend in monthly sales for [Zhou Hui] employees in [2021]?
--SQL1: 
SELECT name,month, sales_amount
FROM employee 
WHERE name = "Zhou Hui"
AND employee.year = 2021
ORDER BY month ASC;

--Query2: How have employee [Zhao Li]'s monthly sales changed over the last year?
--SQL2:
SELECT name,year,month,sales_amount
FROM employee 
WHERE year = YEAR(CURDATE())-1
AND name = "Zhao Li"
GROUP BY month
ORDER BY month ASC;
\end{minted}
\vspace{-15pt}
\caption{In-context examples.}\label{code:examples}
\end{figure}

Utilizing the in-context learning capability of the LLM agent, we integrate the schema information with the exemplary queries and responses, alongside the original user inquiry, to generate an accurate SQL statement that addresses the user's needs, as shown in Figure~\ref{code:SQL}.
\begin{figure}[htbp]
\vspace{-10pt}
\centering
\begin{minted}[breaklines,fontsize=\scriptsize]{SQL}
SELECT name,month, sales_amount
FROM employee 
WHERE name = "Zhou Hui"
AND (month BETWEEN 7 AND 12)
AND employee.year = 2022
ORDER BY month ASC;
\end{minted}
\vspace{-15pt}
\caption{The generated SQL.}\label{code:SQL}
\end{figure}

\begin{figure}
    \vspace{-10pt}
    \centering
    \includegraphics[width=0.98\linewidth]{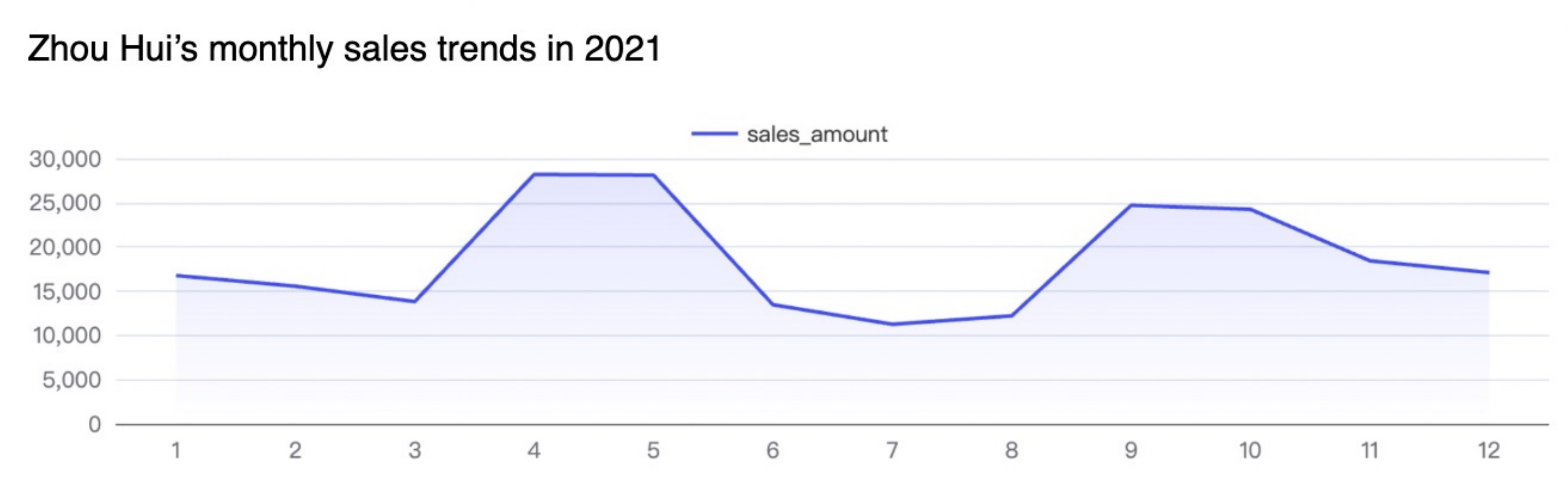}
    \vspace{-15pt}
    \caption{An example of generated line chart.}
    \label{fig:employee_sales}
\end{figure}

\subsection{Chart Generation} Albeit the generation of an accurate SQL statement, presenting the resulting data to users in an intuitive format remains a challenge. To bolster user experience, two chart generation methods were employed, leveraging the in-context capabilities of the LLM agent and external knowledge bases.

The first method involves generating complete eCharts JSON code. For instance, for the real user question concerning the sales trend of employee [Zhou Hui] in [2021], the LLM agent determines that a line chart would best illustrate the desired trend. The final eCharts code, which includes additional interactive features like a title, legend, and tooltips, is crafted through a prompt, with the result displayed in Figure~\ref{fig:employee_sales}.

\begin{figure}
    \vspace{-10pt}
    \centering
    \includegraphics[width=0.98\linewidth]{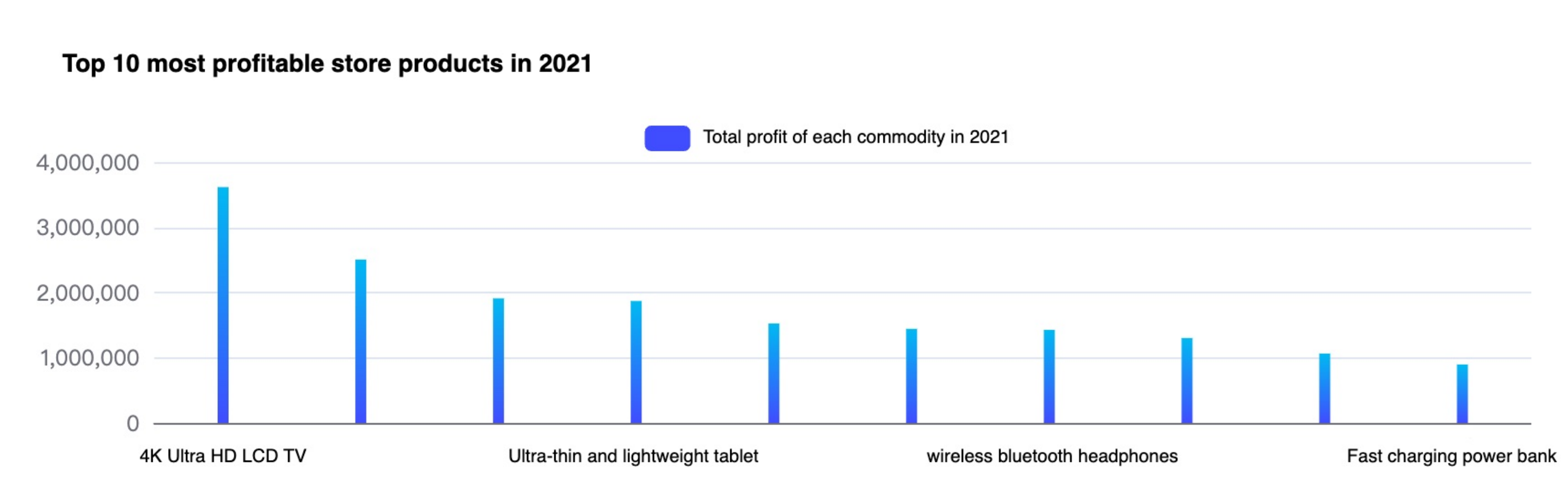}
    \vspace{-15pt}
    \caption{An example of generated bar chart.}
    \label{fig:profit_top10}
\end{figure}

The second method applies rule normalization to expedite the LLM agent's processing time and diminish variation in personalized scenarios. We introduced a simplified prompt structure for constructing basic two-dimensional charts. This is exemplified in the query about the highest total profits among products for March 2022, where we translate SQL results into a list before extracting column names for prompt incorporation, as outlined in Appendix~\ref{sec: axis_checker}. The immediate inference is that product names populate the x-axis and total profits the y-axis, with a bar chart effectively emphasizing the ranking. This refined process enables prompt, adaptable insertion of query results into tailored chart templates, evidenced by the output in Figure~\ref{fig:profit_top10}.

\subsection{Analysis and Summary Capability} While general LLMs are adept with vast textual information, their proficiency in custom and specialized vertical knowledge domains can be limited. To address specialized queries more empathetically and accurately, we have developed a pipeline incorporating prompt engineering and external knowledge bases. For example, in analyzing and predicting time-series data, the \ours{} begins by consulting an external knowledge base for relevant insights into the user's issue. It then assimilates this information, grasps the user's underlying intent, and proceeds with the generation and execution of the corresponding SQL statements to gather the relevant time-series data. It's necessary to call a time-series prediction model for further analysis, with the LLMs for accessing the prediction model, referencing pertinent API documentation from the knowledge base. It then delivers insightful analyses to the user, including trends, periodic behavior, and outlier detection.

\section{The Example of Prompts and Cases}
In this section, we disclose the prompts and actual cases employed in the development of \ours{}.  
\subsection{Prompt: Text Analysis.}\label{sec: text_analysis}
\begin{figure}[htbp]
\vspace{-10pt}
\centering
\begin{minted}[breaklines,fontsize=\scriptsize]{python}
TEXT_ANALYSIS_TEMPLATE = """
Question: {query}
Database Query Result: {result}
Based on the provided question and the results from the database query, please describe the results line by line using appropriate language. The description should cover the full results and provide a concise conclusion, focusing solely on the core issue. There is no need to display any charts, but if the results are related to time, please infer the relevant time range retrospectively. If the query results are empty, return a suitable description without making any assumptions.  
"""
\end{minted}
\vspace{-15pt}
\caption{Text analysis prompt.}\label{code:text_analysis}
\end{figure}

\subsection{integretion.}\label{sec: sql2nl}
\begin{figure}[htbp]
\vspace{-10pt}
\centering
\begin{minted}[breaklines,fontsize=\scriptsize]{python}
SQL2NL_TEMPLATE = """You are now required to generate user questions based on table information and SQL statements: {table_info}
Generate 3 instances of user questions based on table information and SQL statements:
[SQL Statements]: {sql}
[Generated Questions]:"""
\end{minted}
\vspace{-10pt}
\caption{SQL2NL prompt.}\label{code:SQL2NL}
\end{figure}

\subsection{Prompt: Axis Checker.}\label{sec: axis_checker}
\begin{figure}[hbp]
\vspace{-10pt}
\centering
\begin{minted}[breaklines,fontsize=\scriptsize]{python}
AXIS_CHECKER = """Based on the problem and chart description analysis, select the horizontal and vertical coordinates from {column_name}
Keep the original information in column_name
Chart types: line, bar, pie
Based on the table description analysis, select only any one type from the chart types.
Only returns a ```json``` format code
The structure only contains
{{
    "xAxis": "",
    "yAxis": "",
    "type": ""
}}
Table description: {column_desc}"""
\end{minted}
\vspace{-10pt}
\caption{Axis checking prompt.}\label{code:axis}
\end{figure}

\subsection{Prompt: Chart Generation.
\small{[Figure~\ref{code:chart_gen}]}.}
\label{sec: chart_generation}
\begin{figure}[htbp]
\vspace{-10pt}
\centering
\begin{minted}[breaklines,fontsize=\scriptsize]{python}
CHART_GENERATION_TEMPLATE = """
Example: 
Question: In May 23, what are the different types of application single numbers for X?
Chart_type: bar
Answer: [{{'process_type': '2', 'num': '1145'}}, {{'process_type': '5', 'num': '406'}}, {{'process_type': '1', 'num': '505'}}, {{'process_type': '4', 'num': '596'}}, {{'process_type': '0', 'num': '84'}}, {{'process_type': '7', 'num': '33'}}, {{'process_type': '6', 'num': '19'}}]
Output:
The answer has seven sets of data, the chart type is "bar", and a JSON needs to be output.
```json
{{
    "xAxis": {{
        "type": "category",
        "name": "Application form type",
        "data": ["2", "5", "1", "4", "0", "7", "6"]
      }},
      "yAxis": {{
        "type": "value",
        "name": "Application Form quantit"
      }},
      "series": [
        {{
            "data": ["1145", "406", "505", "596", "84", "33", "19"],
            "name": "Application Form of X for May 23",
            "type": "bar"
        }}
      ]
}}
```
Question: {query}
Chart_type: {chart_type}
Answer: {sql_result}
I want you to act like an eCharts builder, an expert in creating meaningful charts.
Completely refer to the above [Example], analyze the data in the answer, and return an ECharts configuration option to present the data results
Output:"""
\end{minted}
\vspace{-10pt}
\caption{Chart generation prompt.}\label{code:chart_gen}
\end{figure}

\subsection{Case: Key Word Missing in Generated SQL.}\label{case:key_word}
\begin{figure}[H]
\vspace{-10pt}
\centering
\begin{minted}[breaklines,fontsize=\scriptsize]{SQL}
-- Question: What is the closure rate of online issues for the Intelligent Office Platform Department in August 2023?
-- pred:
SELECT
  SUM(
    CASE
      WHEN closed_time IS NOT NULL
      AND closed_time > begin_time THEN 1
      ELSE 0
    END
  ) AS closed_count, ...
-- gold:
SELECT
  CONCAT (
       ROUND(
        COUNT(IF (status IN ('closed', 'finished', 'published'), 1, NULL)) / COUNT(*) * 100,
         2
     ),
      '%'
     ) AS close_ratio FROM...
\end{minted}
\vspace{-10pt}
\caption{The missed key word.}\label{code:miss_word}
\end{figure}

\begin{figure}[tbhp]
\centering
\vspace{-2mm}
\includegraphics[width=0.8\linewidth]{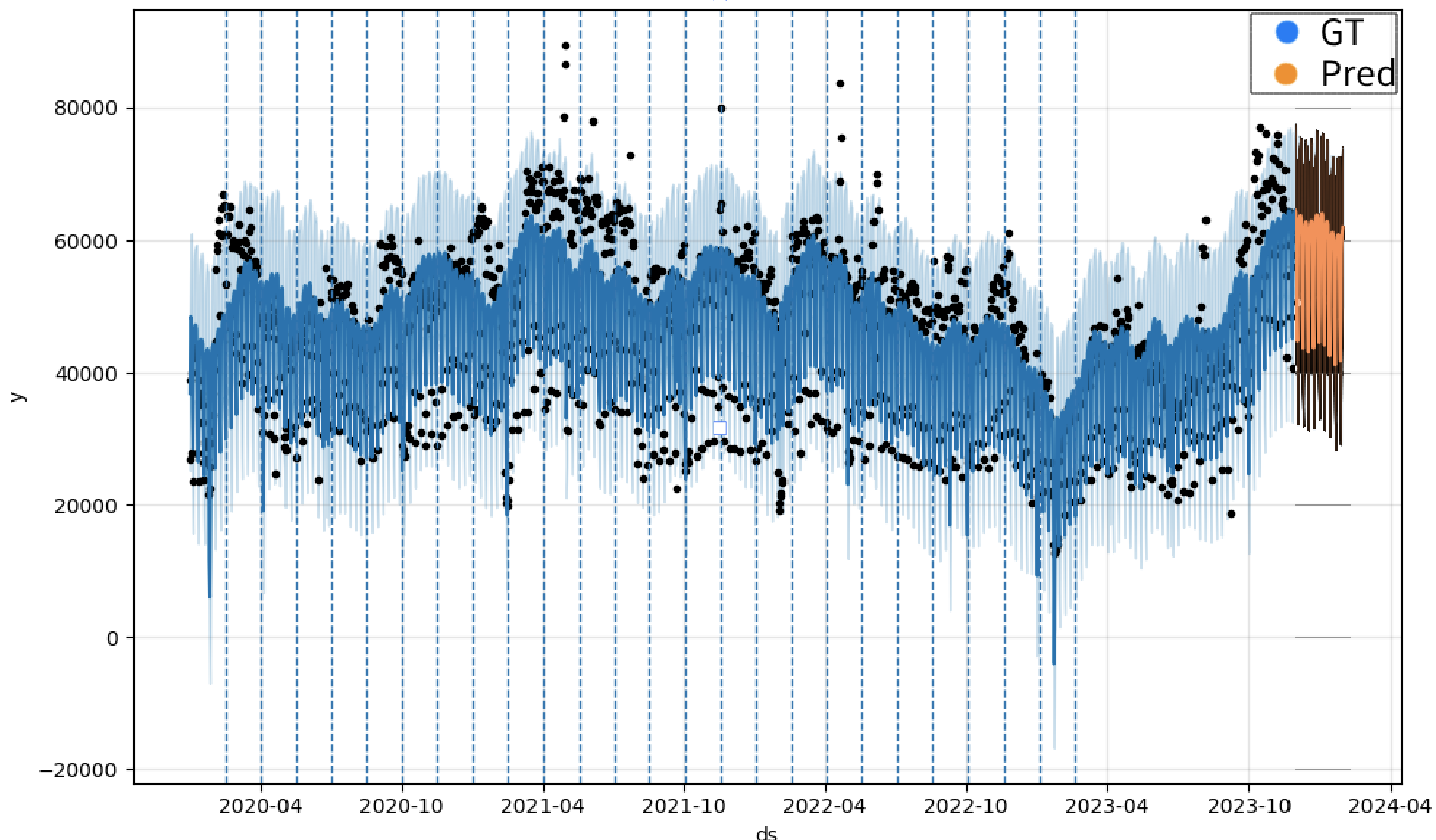}
\vspace{-10pt}
\caption{A forecast result guided by \ours{}.}
\label{fig:prediction}\vspace{-2mm}
\end{figure}

\subsection{Case: Prediction with Prophet \small{[Figure~\ref{fig:prediction}]}.}
The Prophet\footnote{https://github.com/facebook/prophet} forecasting model is known for its intuitive approach to time series data by focusing on components like trend, seasonality, and holidays. Integrating Prophet with LLMs represents a novel area of study, wherein the strengths of both are leveraged for enhanced forecasting and analytical capabilities. Cooperation between Prophet and \ours{} can manifest in data preprocessing and post-analysis. LLMs in \ours{} can assist in curating and interpreting relevant textual data to refine Prophet's inputs and potentially expand its external regressors. Post-prediction, LLMs can translate the numerical outputs into natural language, making the insights more accessible for decision-makers. Additionally, LLMs can support interactive query interfaces, where they convey Prophet's forecasts through conversational AI, enabling users to discuss and digest future trends with ease. While Prophet handles the quantitative predictions, LLMs can enrich the user experience by providing contextual understanding and narrative explanations. The synergy between Prophet and \ours{} does not involve direct computational collaboration but rather a complementary integration where the former provides structured forecasts and the latter enhances the interpretability and application of those forecasts.  

\label{fig:prediction_prophet}

\subsection{Case: View Creation to Simplify SQL.}\label{case:sql_view}
An SQL view is essentially a virtual table created by a query, which provides a way to encapsulate complex SQL operations into a simpler form, acting much like a real table with rows and columns. Views offer various advantages, including simplifying database interactions for users by hiding the complexity of the underlying data operations, enhancing security by limiting data exposure, offering a level of abstraction from database schema changes, and creating a logical representation of data that corresponds with business requirements and user roles. These benefits make views a powerful feature for efficient database management. An example is shown as follows.  
\begin{figure}[H]
\vspace{-10pt}
\centering
\begin{minted}[breaklines,fontsize=\scriptsize]{SQL}
CREATE VIEW table_view
AS
SELECT get_json_object(d2, '$.field1') AS field1
	, get_json_object(d2, '$.field2') AS field2
	, get_json_object(d2, '$.field3') AS field3
	, get_json_object(d2, '$.field4') AS field4
	, get_json_object(d2, '$.field5') AS field5
	, get_json_object(d2, '$.field6') AS field6, stat_date
	, d2, type
FROM table_a
WHERE ((get_json_object(d2, '$.field1') != 'value1'
			AND get_json_object(d2, '$.field2') NOT IN ('id1', 'id2', 'id3'))
		OR (get_json_object(d2, '$.field2') IN ('id1', 'id2', 'id3')
			AND (get_json_object(d2, '$.field3') != 0
				OR get_json_object(d2, '$.field4') IS NOT NULL
				OR get_json_object(d2, '$.field6') = '1'))
		OR (get_json_object(d2, '$.field5') = 'value2'
			AND get_json_object(d2, '$.field6') = '1'))
	AND type = 'type1';
\end{minted}
\vspace{-10pt}
\caption{View creation to simplify SQL.}\label{code:sql_view}
\end{figure}

\section{Supplementary Experimental Results}
\label{sec:feedbacks}
The efficacy of two additional enhancements, which significantly improve both the effectiveness and the user experience of our system, is summarized below.
 

\subsection{Human-to-Machine Feedback} 
\begin{figure}[htbp]
\vspace{-5pt}
\centering
\begin{minipage}{.22\textwidth}
  \centering
  \includegraphics[width=.98\linewidth]{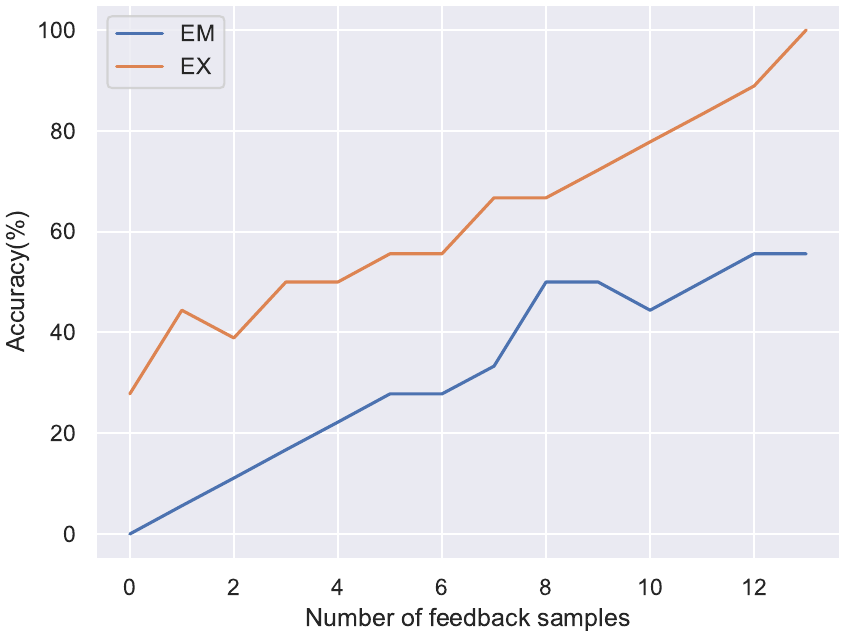}
  \vspace{-10pt}
  \captionof{figure}{The effect of human-to-machine feedback.}
  \label{fig:human_to_machine_feedback}
\end{minipage}%
\hspace{8pt}
\begin{minipage}{.22\textwidth}
  \centering
  \includegraphics[width=.85\linewidth]{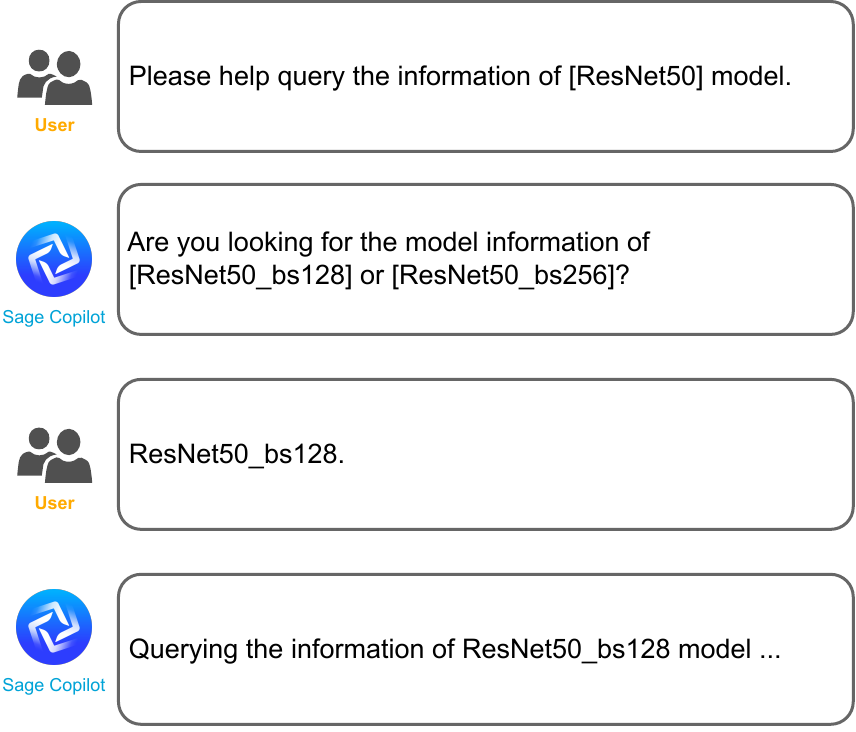}
  \vspace{-10pt}
  \captionof{figure}{An example of machine-to-human feedback.}
  \label{fig:example_of_machine_to_human_feedback}
\end{minipage}
\vspace{-10pt}
\end{figure}

Recognizing the limitations and variability in outputs from Large Language Models (LLMs), it is imperative to establish a feedback loop from humans to machines. This allows for improved precision through active adjustments. In particular, when the system encounters an erroneous case, the incorrect SQL generated in response to a query is manually rectified and incorporated into the memory base to immediately enhance system performance. As depicted in Figure~\ref{fig:human_to_machine_feedback}, implementing this human-to-machine feedback mechanism showcases a marked advancement in output quality.

\subsection{Machine-to-Human Feedback} Traditionally, LLMs deliver responses to queries without regard for result accuracy, which can be problematic. To rectify this, our system implements a machine-to-human feedback process. As illustrated in Figure~\ref{fig:example_of_machine_to_human_feedback}, when there is uncertainty in providing an accurate response, the system solicits human intervention. The response is then formulated using the system's capacity for multi-turn dialogue comprehension, but only after receiving a question that has been clarified by a human operator. This is operationalized by defining a set of clarification parameters alongside acceptable values. If a query involves a clarification parameter and the inferred value does not align with the anticipated range, a request for further clarification is initiated. This process ensures that responses are not only accurate but are also reflective of the user's true intentions, optimizing the system's reliability through iterative machine-to-human feedback.

In the realm of human-assisted feedback, we demonstrated the substantial gain from a human-to-machine feedback mechanism in rectifying erroneous SQL cases, while machine-to-human feedback processes addressed accuracy in responses, ensuring queries were clarified by human intervention when necessary. 




\end{document}